\documentclass[conference]{ieeeconf}
\IEEEoverridecommandlockouts
\pdfoutput=1
\usepackage{amsmath,amssymb,amsfonts}
\usepackage{graphicx}
\usepackage{textcomp}
\usepackage{xcolor}
\usepackage{float, svg, biblatex, tikz, dblfloatfix}
\def\BibTeX{{\rm B\kern-.05em{\sc i\kern-.025em b}\kern-.08em
    T\kern-.1667em\lower.7ex\hbox{E}\kern-.125emX}}
\usetikzlibrary {arrows.meta, angles, quotes}

\newtheorem{defn}{Definition}[section]
\newtheorem{thm}{Theorem}[section]

\begin{document}

\title{\LARGE \bf Autonomous Wheel Loader Navigation Using \\ Goal-Conditioned Actor-Critic MPC
\thanks{This work was supported in part by Academy of Finland (SA) 345517, under SA-NSF joint call on artificial intelligence and wireless communication, and in part by NSF through the RI Grant 2133656. The work is also part of the FEMMa project (2801/31/2021), funded by Business Finland.}
}

\author{Aleksi Mäki-Penttilä\\
\textit{Tampere University}\\
aleksi.maki-penttila@tuni.fi
\and
Naeim Ebrahimi Toulkani\\
\textit{Tampere University}\\
naeim.ebrahimitoulkani@tuni.fi
\and
Reza Ghabcheloo\\
\textit{Tampere University}\\
reza.ghabcheloo@tuni.fi
}

\maketitle

\begin{abstract}
This paper proposes a novel control method for an autonomous wheel loader, enabling time-efficient navigation to an arbitrary goal pose. Unlike prior works which combine high-level trajectory planners with Model Predictive Control (MPC), we directly enhance the planning capabilities of MPC by incorporating a cost function derived from Actor-Critic Reinforcement Learning (RL). Specifically, we first train an RL agent to solve the pose reaching task in simulation, then transfer the learned planning knowledge to an MPC by incorporating the trained neural network critic as both the stage and terminal cost. We show through comprehensive simulations that the resulting MPC inherits the time-efficient behavior of the RL agent, generating trajectories that compare favorably against those found using trajectory optimization. We also deploy our method on a real-world wheel loader, where we demonstrate successful navigation in various scenarios.
\end{abstract}

Autonomous Vehicle Navigation, Optimization and Optimal Control, Motion and Path Planning

\section{Introduction}
Wheel loaders are versatile material-moving machines widely used in industries such as mining, construction and waste management. Their articulated frame steering mechanism enables short radius turns and operation in confined
spaces. However, the high level of skill required to fully utilize these capabilities leads to significant variability in productivity among human operators \cite{frank2012wheel, frank2012increasing}. Furthermore, many tasks performed by wheel loaders are repetitive and take place in extreme conditions, such as intense heat or cold, making automation highly desirable.

Prior work on autonomous wheel loader navigation has frequently employed Model Predictive Control (MPC) due to its planning capabilities \cite{shi2020planning, song2022autonomous}. However, when paired with a prediction horizon suitable for real-time execution, MPC may fail to solve complex planning tasks on its own. Therefore, most works incorporate a separate high-level trajectory planner, and employ MPC as a reference-tracking controller. For instance, \cite{shi2020planning} uses an RRT* based planner with adaptive MPC, while \cite{song2022autonomous} integrates an optimization-based planner with Linear Parameter Varying MPC (LPV-MPC). Although both of these works successfully demonstrate the goal reaching capabilities of their method in obstacle-free simulations, they inherently balance a trade-off between optimality and real-time capability of the high-level trajectory planner.

\begin{figure}[t]
    \centering
    \includesvg[height=16em]{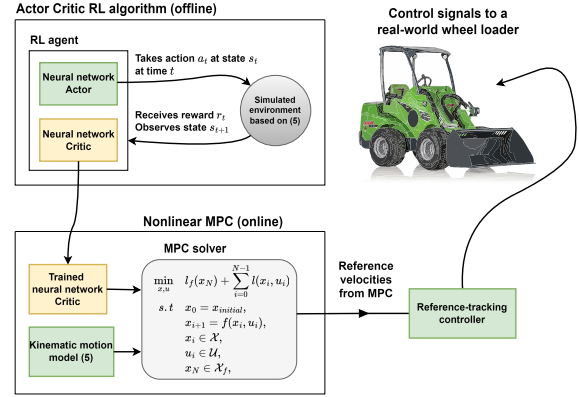}
    \caption{Overview of the proposed control system.}
    \label{fig:control_system}
\end{figure}

Specifically, following outdated trajectories from a non-real-time planner, like the one in \cite{song2022autonomous}, will likely lead to suboptimal performance. This is attributed to inevitable deviations from the planned trajectory due to modeling errors. Real-time capable planners, such as the one in \cite{shi2020planning}, address this issue by providing a constant stream of up-to-date trajectories, thereby minimizing the accumulated deviation from the planned trajectory. However, their real-time execution rate, achieved through sampling or motion discretization, often results in highly suboptimal trajectories.

We propose an alternative control approach which omits the use of a high-level planner, and improves the planning capabilities of the MPC directly. Similar to \cite{reiter2024ac4mpc}, we leverage the fact that a Reinforcement Learning (RL) critic learns to encode all of the information needed to plan approximately optimal behavior. Thus, by integrating a trained RL critic as an MPC cost function, we enable long-term planning even with a prediction horizon suitable for real-time control.

Our method, illustrated in Fig. \ref{fig:control_system}, first trains the Actor-Critic RL agent in simulation using an acceleration limited kinematic model of the wheel loader. The trained critic and kinematic model are then used to formulate a nonlinear MPC problem which generates reference velocities for the front body unit and center joint of the wheel loader. These references are subsequently tracked by a low-level feedback-controller, which commands the steering hydraulics and diesel engine. This cascade control approach ensures precise navigation despite the use of a simplified model in the MPC. The main contributions of this paper are summarized as:
\begin{itemize}
    \item We approximate the time-optimal solution of a wheel loader pose reaching task by solving a Goal-Augmented MDP with a Lyapunov-based RL algorithm, trained in simulation using a constrained kinematic model.
    \item We transfer the planning knowledge of the Actor-Critic RL agent into a nonlinear MPC by formulating suitable terminal and stage costs based on the critic.
    \item We demonstrate the effectiveness of our solution on a real wheel loader, where the RL actor would be unsafe for direct use. A comprehensive study, supported by simulations, shows that our method reaches goal poses faster than a baseline trajectory optimization routine.
\end{itemize}
Compared to prior work on RL wheel loader control \cite{10260481, sardarmehni2023path}, we support a broader range of applications by enabling navigation to an arbitrary goal pose, instead of only executing specific maneuvers. Our approach also enhances safety by enforcing constraints through MPC. Unlike the conventional Actor-Critic MPC in \cite{reiter2024ac4mpc}, we use a Lyapunov-based RL algorithm \cite{wang2023policyoptimizationmethodoptimaltime} to train a critic that serves as a sampling-based Lyapunov function, providing a mean cost stability guarantee. Moreover, we use the critic not only as a terminal cost in our MPC, but also derive a suitable stage cost from it. Lastly, we propose the use of a gradient penalty during RL training to enhance the MPC optimization landscape.

\section{Prior work} \label{sec:priorwork}
\subsection{Actor-Critic Model Predictive Control}
Prior works \cite{reiter2024ac4mpc, romero2024actorcritic} have explored different ways of using RL critics as MPC cost functions. In \cite{romero2024actorcritic}, the authors propose a novel Actor-Critic RL algorithm in which the actor consists of a neural network followed by a differentiable MPC. The actor is trained end-to-end, such that the actor network learns to output the stage-wise coefficients of a quadratic cost for the MPC. However, they are unable to enforce state constraints due to inherent limitations in the differentiable MPC solver \cite{amos2019differentiable}. In contrast, the authors of \cite{reiter2024ac4mpc} train a standard Actor-Critic RL agent and integrate the critic as a terminal cost in a nonlinear MPC problem, which is then solved using a standard Sequential Quadratic Programming (SQP) solver. This allows enforcing state constraints, but due to the highly nonlinear neural network critic, they observe difficulties in solving the underlying optimization problem.

\subsection{Lyapunov neural networks}
Stability analysis using neural Lyapunov functions has received significant attention in the recent years \cite{han2020actorcriticreinforcementlearningcontrol, chang2021stabilizingneuralcontrolusing, wang2023policyoptimizationmethodoptimaltime, wu2023neurallyapunovcontroldiscretetime, yang2024lyapunovstableneuralcontrolstate}. Some approaches, like \cite{wu2023neurallyapunovcontroldiscretetime, yang2024lyapunovstableneuralcontrolstate}, use Counterexample-Guided Inductive Synthesis (CEGIS) to construct Lyapunov functions, which provide strong guarantees through verification methods like Satisfiability Modulo Theory (SMT), but struggle in high-dimensional settings. In contrast, methods based on Actor-Critic RL, such as \cite{han2020actorcriticreinforcementlearningcontrol, chang2021stabilizingneuralcontrolusing, wang2023policyoptimizationmethodoptimaltime}, naturally extend to high-dimensional tasks, but rely on weaker forms of Lyapunov theory, such as almost Lyapunov stability \cite{liu2018lyapunovfunctionsnonlinearsystems} or sampling-based Lyapunov stability, which either tolerate small stability violations or focus on stability in expectation.

\section{Preliminaries} \label{sec:preliminaries}
We consider a goal-augmented Markov Decision Process (MDP) \cite{liu2022goalconditionedreinforcementlearningproblems} with state space $s\in \mathcal{S} \subseteq \mathbb{R}^n$, action space $a \in \mathcal{A} \subseteq \mathbb{R}^n$ and a fixed goal $g \in \mathcal{S}$. The state evolves according to the state transition function $s_{t+1} \sim p(s_{t+1} | s_t, a_t)$, where the action is sampled from a goal-conditioned behaviour policy $a_t \sim \pi(a_t | s_t, g)$. The reward function is defined as $r(s_t, a_t, g): \mathcal{S} \times \mathcal{A} \times \mathcal{S} \rightarrow \mathbb{R}$, and the associated cost is defined as $c(s_t, a_t, g) = -r(s_t, a_t, g)$. Solving the MDP corresponds to finding the policy which minimizes the discounted infinite-horizon cost $J_\pi = \mathbb{E}_\pi \left[ \sum_{t=0}^\infty \gamma^t c(s_t, a_t, g)\ |\ s_0 = s \right]$, where $\gamma \in (0, 1]$ is the a discount factor.

We define the stationary distribution of the state transition\footnote{Equivalent to the closed loop dynamics of a stochastic control system.} as $\mathcal{P}_\pi(s_{t+1} | s_t, g) = \int \pi(a | s_t, g) p(s_{t+1} | s, a) da$. The stationary distribution of the state is subsequently defined as $\mathcal{S}_\pi = \lim_{T \rightarrow \infty} \frac{1}{T} \sum_{t=0}^T \mathcal{T}_\pi(s_t | s_0, g)$, where:
\begin{equation}
    \mathcal{T}_\pi(s_{t+1} | s_0, g) = \int \mathcal{P}_\pi(s_{t+1} | s_t, g) \mathcal{T}_\pi(s_t | s_0, g) ds_t.
\end{equation}

We assume a cost $c(s, a, g)$ defined as a norm between the state and the goal, making it desirable to drive the cost to zero. This enables us to analyze the asymptotic behaviour of the MDP system via the mean cost stability framework \cite{han2020actorcriticreinforcementlearningcontrol}. 

\begin{defn}
\label{def:mean_cost}
(Mean cost stability \cite{han2020actorcriticreinforcementlearningcontrol}). A MDP system is said to be stable in the mean cost under $\pi$ when: 
\begin{equation}
   \lim_{t \rightarrow \infty} \mathbb{E}_{s_t} \left[ c_\pi(s_t, g)\right] = 0
   \label{eq:mean_cost}
\end{equation}
holds for any $s_0 \in \{ s\ |\ c_\pi(s, g) \leq b \}$, where $b \geq 0$ is a constant and $c_\pi(s_t, g) = \mathbb{E}_{a_t \sim \pi(a_t | s_t, g)} \left[ c(s_t, a_t, g) \right]$.
\end{defn}

\begin{thm}
\label{the:mean_cost}(Sampling-based Lyapunov stability \cite{wang2023policyoptimizationmethodoptimaltime}). The mean cost stability of a system can be shown through a function $L(s, g): \mathcal{S} \times \mathcal{S} \rightarrow \mathbb{R}$ when it satisfies the conditions:
\begin{subequations}
\begin{gather}
    k_l c_\pi(s, g) \leq L(s, g) \leq k_u c_\pi(s, g), \label{eq:mc_prop1} \\
    L(s, g) \geq c_\pi(s, g) + \lambda \mathbb{E}_{s' \sim \mathcal{P}_\pi} \left[ L(s', g) \right], \label{eq:mc_prop2} \\
    \begin{split}
        -k\left( \mathbb{E}_{s \sim \mathcal{S}_\pi} \left[ L(s, g) - \lambda \mathbb{E}_{s' \sim \mathcal{P}_\pi} \left[ L(s', g) \right] \right] \right)& \\ \geq \mathbb{E}_{s \sim \mathcal{S}_\pi} \left[ \mathbb{E}_{s' \sim \mathcal{P}_\pi} \left[ L(s', g) \right] - L(s, g) \right], &
    \end{split} \label{eq:mc_prop3}
\end{gather}
\label{eq:mc_props}
\end{subequations}
where $k_l, k_u > 0$ and $k, \lambda \in (0, 1]$ are constants. When the conditions in \eqref{eq:mc_props} are met, then $L(s, g)$ is a valid sampling-based Lyapunov function within $\mathcal{S}_\pi$.
\end{thm}

\section{Methodology} \label{sec:methodology}
We begin with a high-level overview of our approach before detailing its application to our specific problem. Our approach focuses on controlling a nonlinear system:
\begin{equation}
    x_{t+1} = f(x_t, u_t),
\end{equation}
where both the states $x$ and controls $u$ are subject to constraints. Our goal is to drive the system to a goal state $x_g$ and maintain it there indefinitely. To achieve this, we use the description of the nonlinear system to construct a goal-conditioned MDP, where we assign $s = x$, $a = u$, $g = x_g$, and define the state transitions of the MDP as deterministic using $s_{t+1} = f(s_t, a_t)$. We then approximate the optimal solution of the MDP through a Lyapunov-based Actor-Critic RL algorithm \cite{wang2023policyoptimizationmethodoptimaltime}, which trains a policy that stabilizes the MDP system according to Definition \ref{def:mean_cost}. However, while the RL policy is able to solve the control task in simulation, it is unable to take into account actuator limits and other constraints, which complicates real world deployment.

Because of this, we leverage the fact that the RL algorithm also produces a critic that is compliant with Theorem \ref{the:mean_cost}, thereby providing a sampling-based Lyapunov function which can be used to synthesize another controller. Specifically, we integrate the critic as both the stage and terminal cost of a nonlinear MPC, allowing it to inherit the RL agents knowledge on how the system can be stabilized. However, unlike the RL policy, the MPC can enforce constraints, thereby making it better suited for real-world control.

\subsection{Wheel loader kinematic model}
\begin{figure}[H]
\centering
\begin{tikzpicture}[scale=0.75]

% coordinate system
\draw[black, ultra thick, arrows = {-Stealth}] (-4,-2.5) -- (4,-2.5) node[below] {$x$};
\draw[black, ultra thick, arrows = {-Stealth}] (-4,-2.5) -- (-4,4) node[left] {$y$};

% vehicle body:
\draw[black, thick] (-2.83,-0) -- (0,0);
\draw[black, thick] (0,0) -- (2,2);
% front link coordinate
\fill[black] (2,2) circle [radius=0.1];
% center joint coordinate
\fill[black] (0,0) circle [radius=0.1];
% rear link coordinate
\fill[black] (-2.83,0) circle [radius=0.1];

% beta
\draw[black, thick, dotted] (1.5,0) coordinate (A) -- (0,0) coordinate (B) -- (2.47,2.47) coordinate (C);
\pic [pic text=$\beta$, draw=black, angle eccentricity=1.75] {angle = A--B--C};

% front link heading
\draw[black, thick, dotted] (3,3) coordinate (A) -- (2,2) coordinate (B) -- (3.25,2) coordinate (C);
    \draw[black, thick] 
        ([shift={(1.25cm,0cm)}]B) arc[start angle=0,end angle=45,radius=1.25cm] node[midway, right] {$\theta_f$};

% velocities
\draw [blue, ultra thick, arrows = {-Stealth}, rotate around={-45:(2,2)}] (2,2) -- (2, 4.25) node[right] {$v_{f}$};

% wheelbases
\draw[black, thick] (-2.83,-1.5) -- node[below,sloped] {$L_r$} (0,-1.5);
\draw[black, thick] (-2.83,-1.25) -- (-2.83,-1.75);
\draw[black, thick] (0,-1.25) -- (0,-1.75);

\draw[black, thick] (-1.25,1.25) -- node[above,sloped] {$L_f$} (0.75,3.25);
\draw[black, thick, rotate around={45:(-1.25,1.25)}] (-1.25, 1) -- (-1.25,1.5);
\draw[black, thick, rotate around={45:(0.75, 3.25)}] (0.75, 3) -- (0.75,3.5);

% front tire:
\draw[black, thick, rotate around={45:(2,2)}] (1.25,1.7) rectangle (2.75,2.3);

% rear tire:
\draw[black, thick] (-3.58,0.3) rectangle (-2.08,-0.3);

% front body coordinates
\draw[black, thick, dotted] (2,2) -- (-4, 2) node[left] {$y_f$};
\draw[black, thick, dotted] (2,2) -- (2, -2.5) node[below] {$x_f$};

\node[above,font=\small] at (-2.85,0.3) {rear body};
\node[above,font=\small] at (3.125,1) {front body};

\end{tikzpicture}
\caption{Depiction of the wheel loader kinematic model.}
\label{fig:kinematic_model}
\end{figure}
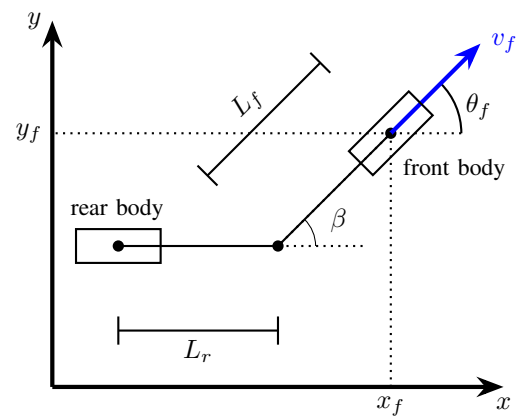

A typical wheel loader is an articulated steering machine, composed of two body units, front and rear as depicted in Fig. \ref{fig:kinematic_model}. Steering is achieved by adjusting the center joint angle $\beta$ between the two body units using a hydraulic actuator. We denote the pose of the front body unit as $(x_f, y_f, \theta_f)$, and its longitudinal velocity as $v_f$. The symbols $L_f$ and $L_r$ are machine specific parameters, which represent the distance from the center joint to the front and rear axle, respectively.

We adapt the wheel loader kinematic model proposed in \cite{5164520} to our purpose. Our model of the system is given by:
\begin{equation}
\label{eq:kinematic_model}
    \dot x = f_c(x, u) =
    \begin{bmatrix}
        v_f \cos(\theta_f) \\
        v_f  \sin(\theta_f) \\
        \frac{L_r \dot \beta + v_f \sin(\beta)}{L_f \cos(\beta) + L_r} \\
        \dot \beta \\
        u_1 \\
        u_2
    \end{bmatrix},
\end{equation}
with $x = [x_f, y_f, \theta_f, \beta, \dot \beta, v_f]^T$ and $u = [u_1, u_2]^T$. The control $u_1$ is the angular acceleration of the center joint, while $u_2$ is the longitudinal acceleration of the front body. 

To ensure the simplified kinematic model remains valid, we enforce the following state constraints:
\begin{equation}
    \begin{bmatrix} 
        -\beta_{max} \\
        -\dot \beta_{max} \\
        -v_{f, max}
        \end{bmatrix}
        \leq
        \begin{bmatrix} 
        \beta \\
        \dot \beta \\
        v_f
        \end{bmatrix}
        \leq
        \begin{bmatrix} 
        \beta_{max} \\
        \dot \beta_{max} \\
        v_{f,max}
    \end{bmatrix}, \label{eq:constraint_x}    
\end{equation}
where the constraint bounds $\beta_{max}$, $\dot \beta_{max}$ and $v_{f, max}$ were determined experimentally using the real wheel loader. 

Finally, we obtain the discrete-time dynamics $x_{t+1} = f(x_t, u_t)$ via the 4th order Explicit Runge-Kutta method:
\begin{equation}
\label{eq:runge_kutta}
x_{t+1} = f(x_t, u_t) = \frac{\Delta t}{6} (k_1 + 2k_2 + 2k_3 + k_4),
\end{equation}
where $\Delta t$ denotes the difference in time between consecutive states $x_t$ and $x_{t+1}$, and $k_1$, $k_2$, $k_3, k_4$ are defined as:
\begin{subequations}
\begin{alignat}{2}
  k_1 &= f_c(x, u), &\ \ k_2 &= f_c(x + \frac{\Delta t}{2} k_1, u), \\ k_3 &= f_c(x + \frac{\Delta t}{2} k_2, u), &\ \ k_4 &= f_c(x + \Delta t k_3, u).
\end{alignat}
\end{subequations}

\subsection{Reinforcement Learning environment design}
We train the RL agent in a simulated environment, which uses the discretized kinematic model \eqref{eq:runge_kutta} for state transitions. We enforce the constraint $\eqref{eq:constraint_x}$ through a clamping mechanism, and set $\dot \beta = 0$ when trying to exceed the joint limit. 

The cost of the environment is defined as a p-norm using the error vector $e = [e_{xy}, e_\theta, \beta, \dot \beta, v_f]^T$ and weights $W$:
\begin{equation}
    c(s, a, g) = ||We||_{0.25},
    \label{eq:env_cost}
\end{equation}
where the position and heading errors, $e_{xy}$ and $e_{\theta}$, are defined based on a goal state $g = [x_g, y_g, \theta_g, 0, 0, 0]^T$ as follows:
\begin{subequations}
    \begin{align}
        e_{xy} &= \sqrt{(x_f - x_g)^2 + (y_f - y_g)^2}, \\
        e_{\theta} &= \arctan_2\left( \sin(\theta_f - \theta_g), \ \cos(\theta - \theta_g) \right).
    \end{align}
\end{subequations}
Importantly, the use of $p=0.25$ in \eqref{eq:env_cost} makes the cost behave in a sparse manner, i.e. giving a nearly constant cost to states which have not converged to the goal. This ensures that the optimal solution of the associated MDP reaches the goal as quickly as possible, since that is the only way to significantly reduce the magnitude of the cost. Other norms, such as the squared Euclidean norm may encourage solutions to initially approach the goal as fast possible, but since the cost values diminish very rapidly when getting closer, there is not enough consistent incentive for time-optimal convergence.

\subsection{Actor and critic neural networks}
We implement our actor and critic, denoted by $\pi_\phi(s, g)$ and $L_\psi\left(s, a, g\right)$, as feedforward neural networks parameterized by $\phi$ and $\psi$. The actor is designed to parameterize a Gaussian action distribution, and the samples from this distribution are bounded by first applying the $\mathrm{tanh}$ function and then scaling by the maximum achievable accelerations $\ddot \beta_{max}$ and $a_{f, max}$. The critic is constructed as:
\begin{equation}
    L_\psi(s, a, g) = Q_\psi(s, a, g) Q_\psi(s, a, g)^T,
\end{equation}
where $Q_\psi(s, a, g)$ is a feedforward neural network parameterized by $\psi$. This ensures positive outputs, which is necessary to satisfy \eqref{eq:mc_prop1} due to the choice of $c(s, a, g)$. We also include an encoder in both networks, which converts all positions into relative positions with respect to the goal and encodes heading angles using sine and cosine values.

% Coordinate Figure -------------------------------------

\subsection{Lyapunov-based Reinforcement Learning}
We train our RL agent using the Adaptive Lyapunov-based Actor-Critic (ALAC) \cite{wang2023policyoptimizationmethodoptimaltime} algorithm, briefly summarized below. In ALAC, the critic is trained to minimize the mean squared error to a target critic $L_{target}\left(s, a, g\right)$:
\begin{equation}
    J_c(\psi) = \mathbb{E}_\mathcal{D} \left[ \frac{1}{2} \left( L_\psi(s, a, g) - L_{target}(s, a, g) \right)^2 \right],
\end{equation}
 where $J_c(\psi)$ denotes the critic loss and $\mathcal{D}$ is an off-policy replay buffer. The target critic is defined as:
 \begin{equation}
 L_{target}(s, a, g) = c(s, a, g) + \gamma \bar{L}_{\bar{\psi}}(s', a', g),
 \end{equation}
 where $\bar{L}_{\bar{\psi}}$ is identical to $L_{\psi}$, except its parameters $\bar{\psi}$ are defined to be an exponentially moving average of $\psi$:
\begin{equation}
    \bar{\psi}_{t+1} = (1-\tau) \bar{\psi}_{t} + \tau \psi_{t},
\end{equation}
with $\tau \in (0, 1]$. The above process corresponds to training the following sampling-based Lyapunov function candidate:
\begin{equation}
    L_\psi(s, g) = \mathbb{E}_\pi \left[ \sum_{t=0}^\infty \gamma^t c(s_t, a_t, g) \ | \ s_0 = s \right],
\end{equation}
which satisfies \eqref{eq:mc_prop1} and \eqref{eq:mc_prop2} \cite{wang2023policyoptimizationmethodoptimaltime}. However, given the difficulties observed in \cite{reiter2024ac4mpc} when optimizing over a neural network critic, we use the following augmented critic loss: 
\begin{equation}
    \label{eq:critic_loss}
    \hat J_c(\psi) = J_c(\psi) + \rho \mathbb{E}_\mathcal{D} \left[ (1 - || \nabla L_\psi(s, a, g) ||_2)^2 \right],
\end{equation}
where the gradient penalty term weighted by $\rho \geq 0$ encourages the critic to be 1-Lipschitz \cite{gulrajani2017improvedtrainingwassersteingans}, thereby providing a smoother optimization landscape for the downstream MPC.

Finally, the actor guides the critic to satisfy the remaining condition (\ref{eq:mc_prop3}) by minimizing the following loss:
\begin{equation}
    J(\phi) = \mathbb{E}_\mathcal{D} \left[ \lambda_e \left( \log \pi_\phi(s, g) + \mathcal{H} \right) + \lambda_l \Delta \mathcal{L}_\psi \right],
    \label{eq:actor_loss}
\end{equation}
where $\lambda_e, \ \lambda_l \in [0, 1]$ are Lagrange variables, $\mathcal{H}$ is the target entropy and $\Delta \mathcal{L}_\psi$ signifies the sample violation of (\ref{eq:mc_prop3}):
\begin{equation}
\begin{split}
    \Delta \mathcal{L}_\psi &= L_\psi\left(s', \pi_\phi(s', g)\right) - L_\psi(s, a, g) \\ & + k\left( L_\psi(s, a, g) - \lambda L_\psi\left(s_{t+1}, \pi_\phi(s', g)\right) \right).
\end{split}
\end{equation}
During training, the Lagrange variables $\lambda_e$ and $\lambda_l$ are adjusted to maximize their corresponding losses:
\begin{subequations}
\begin{align}
    J(\lambda_e) &= \lambda_e \mathbb{E}_\mathcal{D} \left[ \log \pi_\phi(s, g) + \mathcal{H} \right], \\
    J(\lambda_l) &= \lambda_l \mathbb{E}_{\mathcal{D}} \left[ \Delta \mathcal{L}_\psi \right].
\end{align}
\end{subequations}
Intuitively, $\lambda_e$ converges to 0 when we satisfy the minimum entropy constraint $-\log \pi(s, g)_\phi \geq \mathcal{H}$, which has been shown to improve exploration and robustness \cite{eysenbach2022maximumentropyrlprovably}. Similarly, $\lambda_l$ trends towards 0 when (\ref{eq:mc_prop3}) is satisfied for certain $k$ and $\lambda$. Consequently, when $\lambda_l < 1$ then $L_\psi$ is a valid sampling-based Lyapunov function for at least $(s, g) \in \mathcal{D}$. After each training step the parameters $k = 1 - \lambda_l$ and $\lambda = \min(\lambda_l, \gamma)$ self-adjust to prevent premature convergence.
    
\subsection{Model Predictive Control problem formulation}
Our nonlinear MPC, to be described below, is formulated as the following multiple-shooting optimization problem: % Our MPC is stated below and solved using a dual-shooting optimization problem. 
\begin{subequations}
\label{eq:optimization_problem}
\begin{align}
\min_{x, u} \ & l_f(x_N, g) + \sum_{n=0}^{N-1} l(x_n, u_n, g), && \\
s.t \ \ & x_0 = x_{initial}, && \\
& x_{i+1} = f(x_i, u_i), && i = 0, .., N-1 \\
& \eqref{eq:constraint_x}, \eqref{eq:obs_constraint}, && i = 1, .., N \label{eq:mpc_constrain_x} \\
& (\ref{eq:constraint_u}), && i = 0, .., N-1 \label{eq:mpc_constraint_u}
\end{align}
\end{subequations}
where $N$ is the prediction horizon. The predictive model of the MPC is the discretized kinematic model \eqref{eq:runge_kutta}, with state $x = [x_f, y_f, \theta_f, \beta, \dot \beta, v_f ]^T$ and controls $u = [u_1, u_2]^T$, which are the time derivatives of the last two state variables. However, the inputs to our low-level controller are velocities and not accelerations, that is, they are $(\dot \beta_{cmd}, v_{f,cmd})$. Therefore, at each sample time, we select $\dot \beta$ and $v_f$ from the first solution state $x_1$ and send them to the low-level controller, which subsequently commands the real-world machine. 

For all stages of MPC, we enforce the state constraint in \eqref{eq:constraint_x} along with the following constraint on the controls:
\begin{equation}
    \begin{bmatrix} 
    -\Ddot{\beta}_{max} \\
    -a_{f,{max}}
    \end{bmatrix}
    \leq
    \begin{bmatrix} 
    u_1 \\
    u_2
    \end{bmatrix}
    \leq
    \begin{bmatrix} 
    \Ddot{\beta}_{max} \\
    a_{f,{max}}
    \end{bmatrix}.
    \label{eq:constraint_u}
\end{equation} 
We also take into account the presence of circular obstacles by enforcing the following constraint at each stage:
\begin{equation}
    \label{eq:obs_constraint}
    (x_f - x_o^i)^2 + (y_f - y_o^i)^2 \geq (R_o^i)^2,
\end{equation}
where obstacle $i$ is centered at $(x_o^i, y_o^i)$ and has radius $R_o^i$. This formulation assumes that the radius of the obstacle is expaned to include the radius of the wheel loader front body. 

Similar to \cite{reiter2024ac4mpc}, we define the MPC terminal cost $l_f(x_N, g)$ as the RL critic, with the actions replaced by a zero vector: 
\begin{equation}
    l_f(x_N, g) = L_\psi\left( x_N, \mathbf{0}_{2\times1}, g \right).
\end{equation} 
However, we deviate from \cite{reiter2024ac4mpc} in defining the stage cost. The stage cost of MPC problem for stage $n$ is defined as:
\begin{equation}
    l(x_n, u_n, g) = \Delta t \tilde{L}(x_n, u_n, g),
\end{equation}
where $\tilde{L}(x_n, u_n, g)$ is a second order Taylor approximation of the RL critic around the previous MPC solution $(x^*, u^*)$:
\begin{equation}
\begin{split}
    \tilde{L}(x_n, u_n, g) = \ &\frac{\partial L(z^*_{n+1}, g)}{\partial z_n} (z^*_{n+1} - z_n) \\ \ + \ & 0.5 \frac{\partial^2 L(z^*_{n+1}, g)}{\partial z_n^2} (z^*_{n+1} - z_n)^2,
    \end{split}
\end{equation}
with $L(z^*_{n+1}, g) = L_\psi \left(x^*_{n+1}, u^*_{n+1}, g\right)$ and the definitions:
\begin{equation}
    z^*_{n+1} = \begin{bmatrix}x^*_{n+1} \\ u^*_{n+1}\end{bmatrix}, \quad z_n = \begin{bmatrix}x_n \\ u_n\end{bmatrix}.
\end{equation}
Without the stage cost defined above, we observed indecisive behaviour from the machine during the non-terminal stages. However, the additional stage cost derived from the critic increases the computational complexity significantly, which was alleviated by the use of a Taylor approximation. 

\newpage

\section{Experiments And Results} \label{sec:experiments}

\subsection{Experimental setup}
We train our RL agent using PyTorch \cite{gardner2021gpytorchblackboxmatrixmatrixgaussian}, and base our implementation on stable-baslines3 \cite{stable-baselines3}. The actor and critic networks consist of layers with $(48, 96, 144, 96, 48)$ hidden units, employing the $\mathrm{SoftPlus}$ activation function. The RL agent is trained until convergence to $\lambda_l = 0.8$, indicating that the critic is a sampling-based-Lyapunov function.

We formulate the MPC problem in \eqref{eq:optimization_problem} using CasADi \cite{Andersson2019} and Acados \cite{acados}, using L4CasADi \cite{salzmann2023learningcasadidatadrivenmodels, Salzmann_2023} to integrate the neural network critic. The resulting optimization problem is solved using a SQP-RTI scheme, where we utilize the HPIPM \cite{frison2020hpipmhighperformancequadraticprogramming} QP solver. The solver is warm started using the solution from the previous iteration. During our experiments we use $N=10$ and $\Delta t = 0.2 \ s$, which were chosen to balance the trade-off between real-time capability, length of the prediction horizon and discretization error.

\begin{figure}[t]
    \vspace{5pt}
    \centering
    \includegraphics[height=150pt]{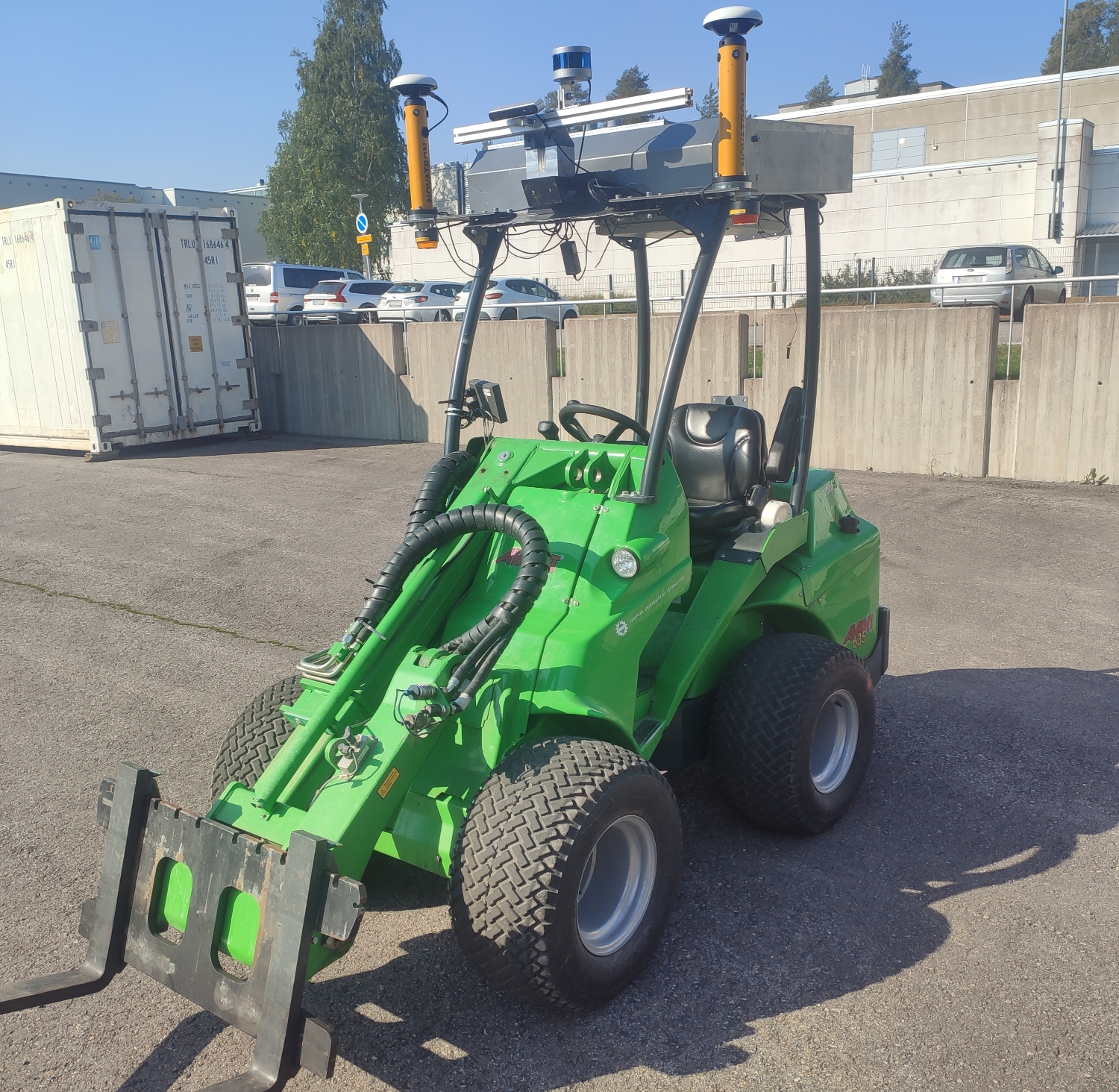}
    \caption{The Avant 635 wheel loader used in our experiments.}
    \label{fig:avant}
\end{figure}

We conduct the real-world experiments using a small Avant 635 wheel loader, shown in Fig. \ref{fig:avant}, which has been retrofitted for autonomous operation. During these experiments, all computations are performed on an onboard NVIDIA Jetson AGX Orin developer kit. The real-world actuators have an input delay of roughly 200 ms, which we compensate by propagating the MPC initial state forwards by the same amount using \eqref{eq:kinematic_model}. The motion constraints used for RL, MPC and the baseline are given in Table \ref{tab:parameters}.

\subsection{Baseline trajectory optimization}
The time-efficacy of our method is evaluated through comparisons to the following trajectory optimization procedure:
\begin{subequations}
\begin{align}
\min_{x, u} \ & \int_{t=0}^{T} \sqrt[4]{e(t)} + \beta(t)^2 + \ddot \beta(t)^2 + a_f(t)^2 \ dt, \\
s.t \quad & x(0) = x_{initial}, \ x(T) = x_{goal}, \\
& \dot x = f_c(x(t), u(t)), \\
& \eqref{eq:constraint_x},\ \eqref{eq:constraint_u},\ \eqref{eq:obs_constraint}, 
\end{align}\label{eq:traj_opt}%
\end{subequations}
where the dynamics $f_c(x(t), u(t))$ are given by the kinematic model \eqref{eq:kinematic_model}, and the error term $e(t)$, designed to encourage convergence to the goal before $t=T$, is defined as:
\begin{equation}
\begin{split}
    e(t) = \ &c_1(x_g-x_f(t))^2 + c_1(y_g-y_f(t))^2 
    \\ +\ &c_2(1-\cos(\theta_g-\theta_f(t)))^2 + \epsilon,
\end{split}
\end{equation}
where $\epsilon, c_1, c_2 > 0$ are constants. We formulate \eqref{eq:traj_opt} using Casadi \cite{Andersson2019} and discretize it using direct collocation with a sampling time of 200 ms. The resulting optimization problem is solved using IPOPT \cite{wachter2006implementation} with $T=25\ s$. This baseline aims to reflect the ideal performance achievable with a traditional control approach, disregarding tracking errors.

\renewcommand{\arraystretch}{1.33}
\begin{table}[H]
    \centering
    \begin{tabular}{|c|c|}
        \hline
        $L_f = L_r$ & 0.6 m \\
        \hline
        $\beta_{max}$ & 40 deg \\
        \hline
        $v_{f, max}$ & 1 m/s \\
        \hline
    \end{tabular}
    \begin{tabular}{|c|c|}
        \hline
        $\dot \beta_{max}$ & 33 deg/s \\
        \hline
        $\ddot \beta_{max}$ & 33 deg/$s^2$ \\
        \hline
        $a_{f_{max}}$ & 1 m/$s^2$ \\
        \hline
    \end{tabular}
    \caption{Motion constraints used during the experiments.}
    \label{tab:parameters}
\end{table}

\begin{figure}[b]
    \centering    
    \includesvg[height=160pt]{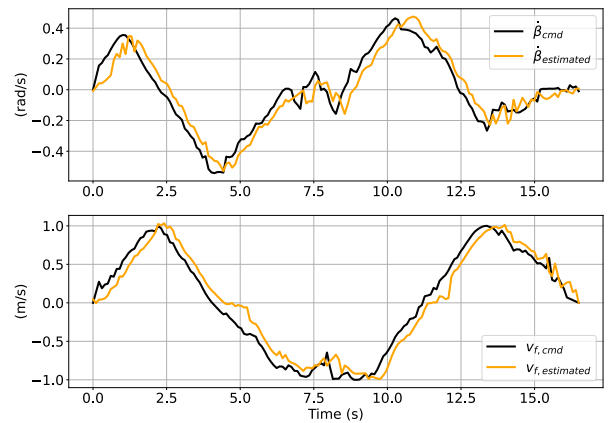}
    \caption{Commanded and estimated velocities for scenario (b).}
    \label{fig:tracking_performance}
 \end{figure}

\begin{figure*}[]
    \vspace{5pt}
    \centering
    \includesvg[inkscapelatex=false, width=450pt]{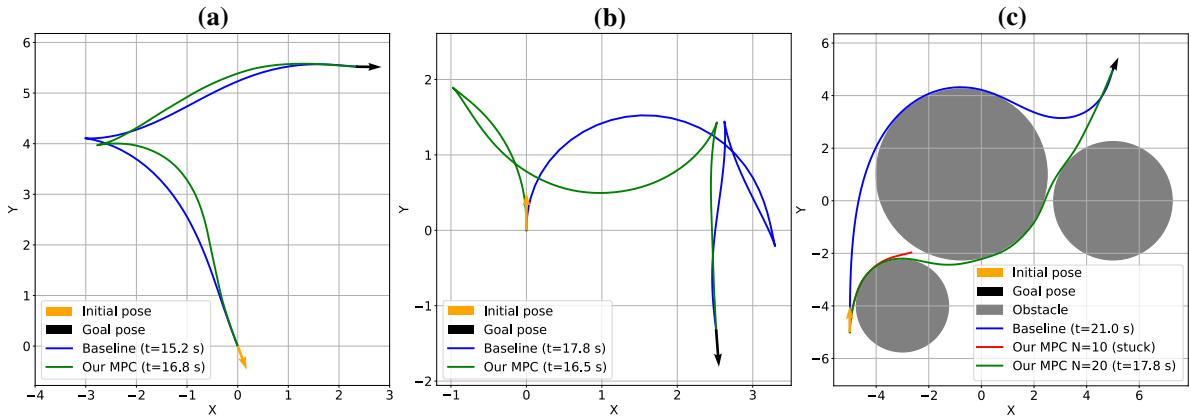}
    \caption{The three highlighted scenarios. (a) Short loading cycle (b) Compact 180-degree turn (c) Navigation through multiple obstacles. Scenarios (a) and (b) were evaluated in the real world, while scenario (c) was conducted using simulations. The time $t$ signifies the first time instant when $||x - g|| < 0.1$. For scenario (c) we illustrate trajectories for both $N=10$ and $N=20$ to highlight the dependence on a sufficiently long prediction horizon.}
    \label{fig:detail_scenarios}
\end{figure*}

\subsection{Highlighted scenarios}
We analyze three specific scenarios in detail: (a) a short loading cycle, (b) a compact 180-degree turn, and (c) navigation through multiple obstacles. Scenarios (a) and (b) were evaluated using the real wheel loader. Scenario (c) had to be simulated using the kinematic model \eqref{eq:kinematic_model}, since successful execution required a prediction horizon of $N=20$, which was too long for real-time execution on the Jetson.

As shown in Figure \ref{fig:detail_scenarios}, the Actor-Critic MPC closely matches the baseline's convergence time in scenario (a) and slightly outperforms it in the more challenging scenario (b). The velocity tracking error for scenarios (a) and (b) is illustrated in Figure \ref{fig:tracking_performance}, which highlights that the Actor-Critic MPC achieves this competitive performance despite the presence of minor actuator delays and velocity tracking errors. The difference in execution time is also significant: the baseline trajectory optimization takes over 5 seconds on a desktop AMD Ryzen 3900x CPU, while each iteration of the Actor-Critic MPC problem is solved in less than 100 milliseconds on the much weaker CPU of the Jetson. However, using $N=20$ for scenario (c) increases the MPC solve time to 200-300 milliseconds, which prevents successful real-world use in obstacle-rich environments.

\subsection{Simulation study}
\begin{figure}[H]
    \centering
    \includesvg[inkscapelatex=false, height=170pt]{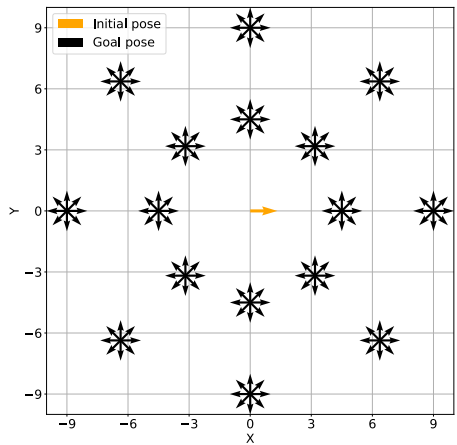}
    \caption{The scenarios used in our supplementary analysis. Each black arrow represents a goal pose, while the orange arrow depicts the initial pose.}
    \label{fig:sim_scenarios}
\end{figure}

\begin{figure}[H]
    \centering
    \includesvg[height=140pt]{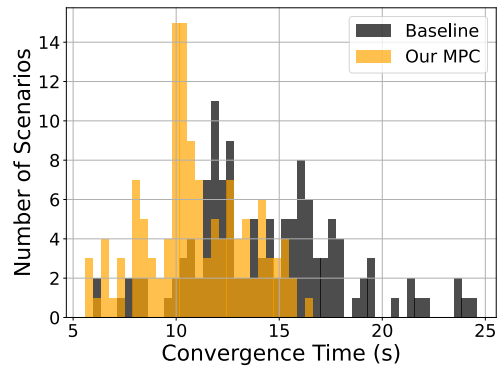}
    \caption{Histogram of convergence times for the 128 simulated scenarios.}
    \label{fig:time_hist}
\end{figure}

To provide a more comprehensive analysis, we use \eqref{eq:kinematic_model} to simulate 128 additional pose reaching scenarios, which are shown in Fig. \ref{fig:sim_scenarios}. We apply our MPC to these scenarios and measure the time taken to converge within $|| x - g || < 0.1$. These convergence times are then compared to those computed for the baseline. The results are drawn to a histogram in Fig. \ref{fig:time_hist} and key statistics are listed in Table \ref{tab:time_table}. Notably, our MPC successfully completes all scenarios, converging 23.80 $\%$ faster on average than the baseline. This advantage becomes evident now that a broader set of scenarios are evaluated, and the real-world modeling errors are eliminated.

\begin{table}[H]
    \centering
    \renewcommand{\arraystretch}{1.3}
    \setlength{\tabcolsep}{12pt}
    
    \begin{tabular}{|l|c|c|}
    \hline
    \textbf{Metric} & \textbf{Mean $\pm$ std} & \textbf{Median} \\
    \hline
    Convergence time of baseline & (14.33 $\pm$ 3.72) s & 13.90 s \\ 
    Convergence time of MPC & (10.92 $\pm$ 2.45) s & 10.60 s \\ 
    \hline
    \textbf{Improvement over baseline} & 23.80 \% & 23.74 \% \\ 
    \hline
    \end{tabular}
    
    \caption{Convergence time statistics for the 128 simulated scenarios.}
    \label{tab:time_table}
\end{table}

\newpage

\section{Conclusion and future work} \label{sec:conclusion}
 We present a novel control scheme based on the Actor-Critic MPC framework, and apply it to solve a wheel loader pose reaching task in a time-efficient manner. Our simulated experiments show that the trajectories generated by our method compare favorably against those found using nonlinear trajectory optimization. Our approach also demonstrates some robustness by succeeding in real-world tests despite the presence of actuator delay and velocity tracking errors. 
 
However, we encounter difficulties in maintaining a real-time control rate when obstacle avoidance is required. Additionally, the MPC occasionally fails to solve certain obstacle-free scenarios in the real world, which we attribute to two main factors: the large number of local minima in the optimization problem, due to the highly nonlinear MPC cost, and the significant actuator delay on our experimental platform, which complicates the control task.
 
Although our empirical experiments indicate that using a Lyapunov-based RL algorithm along with a gradient penalty is advantageous, their significance needs to be studied more rigorously in future work. Additionally, exploring Control Barrier Functions (CBFs) to shorten the required prediction horizon for obstacle avoidance, similar to \cite{zeng2021safetycriticalmodelpredictivecontrol, abdi2024model}, is another promising direction for future research.

\printbibliography[heading=bibintoc]

@inproceedings{frank2012wheel,
  title={On wheel loader fuel efficiency difference due to operator behaviour distribution},
  author={Frank, Bobbie and Skogh, Lennart and Alak{\"u}la, Mats},
  booktitle={2nd International Commercial Vehicle Technology Symposium, CVT},
  pages={1--18},
  year={2012}
}

@inproceedings{frank2012increasing,
  title={On increasing fuel efficiency by operator assistant systems in a wheel loader},
  author={Frank, Bobbie and Skogh, Lennart and Filla, Reno and Fr{\"o}berg, Anders and Alak{\"u}la, Mats},
  booktitle={International conference on advanced vehicle technologies and integration (VTI 2012), Changchun, China},
  year={2012}
}

@inproceedings{5164520,
  title={Modeling and motion control of an articulated-frame-steering hydraulic mobile machine},
  author={Ghabcheloo, Reza and Hyvonen, Mika},
  booktitle={2009 17th Mediterranean Conference on Control and Automation},
  pages={92--97},
  year={2009},
  organization={IEEE}
}

@article{shi2020planning,
  title={Planning the trajectory of an autonomous wheel loader and tracking its trajectory via adaptive model predictive control},
  author={Shi, Junren and Sun, Dongye and Qin, Datong and Hu, Minghui and Kan, Yingzhe and Ma, Ke and Chen, Ruibo},
  journal={Robotics and Autonomous Systems},
  volume={131},
  pages={103570},
  year={2020},
  publisher={Elsevier}
}

@inproceedings{song2022autonomous,
  title={Autonomous wheel loader trajectory tracking control using lpv-mpc},
  author={Song, Ruitao and Ye, Zhixian and Wang, Liyang and He, Tianyi and Zhang, Liangjun},
  booktitle={2022 American Control Conference (ACC)},
  pages={2063--2069},
  year={2022},
  organization={IEEE}
}

@inproceedings{10260481,
  title={Autonomous navigation of wheel loaders using task decomposition and reinforcement learning},
  author={Borngrund, Carl and Bodin, Ulf and Sandin, Fredrik and Andreasson, Henrik},
  booktitle={2023 IEEE 19th International Conference on Automation Science and Engineering (CASE)},
  pages={1--8},
  year={2023},
  organization={IEEE}
}

@article{sardarmehni2023path,
  title={Path Planning and Energy Optimization in Optimal Control of Autonomous Wheel Loaders Using Reinforcement Learning},
  author={Sardarmehni, Tohid and Song, Xingyong},
  journal={IEEE Transactions on Vehicular Technology},
  volume={72},
  number={8},
  pages={9821--9834},
  year={2023},
  publisher={IEEE}
}

@article{amos2019differentiable,
  title={Differentiable mpc for end-to-end planning and control},
  author={Amos, Brandon and Jimenez, Ivan and Sacks, Jacob and Boots, Byron and Kolter, J Zico},
  journal={Advances in neural information processing systems},
  volume={31},
  year={2018}
}

@inproceedings{romero2024actorcritic,
  title={Actor-critic model predictive control},
  author={Romero, Angel and Song, Yunlong and Scaramuzza, Davide},
  booktitle={2024 IEEE International Conference on Robotics and Automation (ICRA)},
  pages={14777--14784},
  year={2024},
  organization={IEEE}
}

@article{reiter2024ac4mpc,
  title={AC4MPC: Actor-Critic Reinforcement Learning for Nonlinear Model Predictive Control},
  author={Reiter, Rudolf and Ghezzi, Andrea and Baumg{\"a}rtner, Katrin and Hoffmann, Jasper and McAllister, Robert D and Diehl, Moritz},
  journal={arXiv preprint arXiv:2406.03995},
  year={2024}
}

@article{liu2018lyapunovfunctionsnonlinearsystems,
  title={Almost Lyapunov functions for nonlinear systems},
  author={Liu, Shenyu and Liberzon, Daniel and Zharnitsky, Vadim},
  journal={Automatica},
  volume={113},
  pages={108758},
  year={2020},
  publisher={Elsevier}
}

@article{han2020actorcriticreinforcementlearningcontrol,
  title={Actor-critic reinforcement learning for control with stability guarantee},
  author={Han, Minghao and Zhang, Lixian and Wang, Jun and Pan, Wei},
  journal={IEEE Robotics and Automation Letters},
  volume={5},
  number={4},
  pages={6217--6224},
  year={2020},
  publisher={IEEE}
}

@inproceedings{chang2021stabilizingneuralcontrolusing,
  title={Stabilizing neural control using self-learned almost lyapunov critics},
  author={Chang, Ya-Chien and Gao, Sicun},
  booktitle={2021 IEEE International Conference on Robotics and Automation (ICRA)},
  pages={1803--1809},
  year={2021},
  organization={IEEE}
}

@inproceedings{wang2023policyoptimizationmethodoptimaltime,
  title={A Policy Optimization Method Towards Optimal-time Stability},
  author={Wang, Shengjie and Fengb, Lan and Zheng, Xiang and Cao, Yuxue and Oseni, Oluwatosin OluwaPelumi and Xu, Haotian and Zhang, Tao and Gao, Yang},
  booktitle={Conference on Robot Learning},
  pages={1154--1182},
  year={2023},
  organization={PMLR}
}

@article{wu2023neurallyapunovcontroldiscretetime,
  title={Neural lyapunov control for discrete-time systems},
  author={Wu, Junlin and Clark, Andrew and Kantaros, Yiannis and Vorobeychik, Yevgeniy},
  journal={Advances in neural information processing systems},
  volume={36},
  pages={2939--2955},
  year={2023}
}

@misc{yang2024lyapunovstableneuralcontrolstate,
      title={Lyapunov-stable Neural Control for State and Output Feedback: A Novel Formulation}, 
      author={Lujie Yang and Hongkai Dai and Zhouxing Shi and Cho-Jui Hsieh and Russ Tedrake and Huan Zhang},
      year={2024},
      eprint={2404.07956},
      archivePrefix={arXiv},
      primaryClass={cs.LG},
}

@article{gulrajani2017improvedtrainingwassersteingans,
  title={Improved training of wasserstein gans},
  author={Gulrajani, Ishaan and Ahmed, Faruk and Arjovsky, Martin and Dumoulin, Vincent and Courville, Aaron C},
  journal={Advances in neural information processing systems},
  volume={30},
  year={2017}
}

@article{stable-baselines3,
  title={Stable-baselines3: Reliable reinforcement learning implementations},
  author={Raffin, Antonin and Hill, Ashley and Gleave, Adam and Kanervisto, Anssi and Ernestus, Maximilian and Dormann, Noah},
  journal={Journal of Machine Learning Research},
  volume={22},
  number={268},
  pages={1--8},
  year={2021}
}

@article{eysenbach2022maximumentropyrlprovably,
  title={Maximum entropy RL (provably) solves some robust RL problems},
  author={Eysenbach, Benjamin and Levine, Sergey},
  journal={arXiv preprint arXiv:2103.06257},
  year={2021}
}

@article{liu2022goalconditionedreinforcementlearningproblems,
  title={Goal-conditioned reinforcement learning: Problems and solutions},
  author={Liu, Minghuan and Zhu, Menghui and Zhang, Weinan},
  journal={arXiv preprint arXiv:2201.08299},
  year={2022}
}

@article{acados,
  title={acados—a modular open-source framework for fast embedded optimal control},
  author={Verschueren, Robin and Frison, Gianluca and Kouzoupis, Dimitris and Frey, Jonathan and Duijkeren, Niels van and Zanelli, Andrea and Novoselnik, Branimir and Albin, Thivaharan and Quirynen, Rien and Diehl, Moritz},
  journal={Mathematical Programming Computation},
  volume={14},
  number={1},
  pages={147--183},
  year={2022},
  publisher={Springer}
}

@article{Salzmann_2023,
  title={Real-time neural MPC: Deep learning model predictive control for quadrotors and agile robotic platforms},
  author={Salzmann, Tim and Kaufmann, Elia and Arrizabalaga, Jon and Pavone, Marco and Scaramuzza, Davide and Ryll, Markus},
  journal={IEEE Robotics and Automation Letters},
  volume={8},
  number={4},
  pages={2397--2404},
  year={2023},
  publisher={IEEE}
}

@inproceedings{salzmann2023learningcasadidatadrivenmodels,
  title={Learning for casadi: Data-driven models in numerical optimization},
  author={Salzmann, Tim and Arrizabalaga, Jon and Andersson, Joel and Pavone, Marco and Ryll, Markus},
  booktitle={6th Annual Learning for Dynamics \& Control Conference},
  pages={541--553},
  year={2024},
  organization={PMLR}
}

@Article{Andersson2019,
  author = {Joel A E Andersson and Joris Gillis and Greg Horn
            and James B Rawlings and Moritz Diehl},
  title = {{CasADi} -- {A} software framework for nonlinear optimization
           and optimal control},
  journal = {Mathematical Programming Computation},
  volume = {11},
  number = {1},
  pages = {1--36},
  year = {2019},
  publisher = {Springer},
}

@inproceedings{zeng2021safetycriticalmodelpredictivecontrol,
  title={Safety-critical model predictive control with discrete-time control barrier function},
  author={Zeng, Jun and Zhang, Bike and Sreenath, Koushil},
  booktitle={2021 American Control Conference (ACC)},
  pages={3882--3889},
  year={2021},
  organization={IEEE}
}

@article{gardner2021gpytorchblackboxmatrixmatrixgaussian,
  title={Gpytorch: Blackbox matrix-matrix gaussian process inference with gpu acceleration},
  author={Gardner, Jacob and Pleiss, Geoff and Weinberger, Kilian Q and Bindel, David and Wilson, Andrew G},
  journal={Advances in neural information processing systems},
  volume={31},
  year={2018}
}

@article{wachter2006implementation,
  title={On the implementation of an interior-point filter line-search algorithm for large-scale nonlinear programming},
  author={W{\"a}chter, Andreas and Biegler, Lorenz T},
  journal={Mathematical programming},
  volume={106},
  pages={25--57},
  year={2006},
  publisher={Springer}
}

@article{frison2020hpipmhighperformancequadraticprogramming,
  title={HPIPM: a high-performance quadratic programming framework for model predictive control},
  author={Frison, Gianluca and Diehl, Moritz},
  journal={IFAC-PapersOnLine},
  volume={53},
  number={2},
  pages={6563--6569},
  year={2020},
  publisher={Elsevier}
}

@inproceedings{abdi2024model,
  title={Model Predictive Control Barrier Functions: Guaranteed Safety with Reduced Conservatism and Shortened Horizon},
  author={Abdi, Hossein and Zhao, Pan and Hovakimyan, Naira and Ghabcheloo, Reza},
  booktitle={2024 American Control Conference (ACC)},
  pages={1652--1657},
  year={2024},
  organization={IEEE}
}

\end{document}